\documentclass[letterpaper]{article} % DO NOT CHANGE THIS
\usepackage{aaai25}  % DO NOT CHANGE THIS
\usepackage{times}  % DO NOT CHANGE THIS
\usepackage{helvet}  % DO NOT CHANGE THIS
\usepackage{courier}  % DO NOT CHANGE THIS
\usepackage[hyphens]{url}  % DO NOT CHANGE THIS
\usepackage{graphicx} % DO NOT CHANGE THIS
\usepackage{natbib}  % DO NOT CHANGE THIS AND DO NOT ADD ANY OPTIONS TO IT
\usepackage{caption} % DO NOT CHANGE THIS AND DO NOT ADD ANY OPTIONS TO IT
\usepackage{algorithm}
\usepackage{algorithmic}

\usepackage{newfloat}
\usepackage{listings}
\DeclareCaptionStyle{ruled}{labelfont=normalfont,labelsep=colon,strut=off} % DO NOT CHANGE THIS
\floatstyle{ruled}
\newfloat{listing}{tb}{lst}{}
\floatname{listing}{Listing}
\nocopyright 
\newcommand{\prb}{\mathds{P}}

\newcommand{\U}{\mathcal{U}}
\newcommand{\Layer}{\mathcal{L}}
\newcommand{\St}{\mathcal{S}}
\newcommand{\A}{\mathcal{A}}

\newcommand{\R}{\mathbb{R}}
\newcommand{\K}{\mathcal{K}}
\usepackage{color}

\newcommand{\argmin}{\arg\!\min}
\newcommand{\argmax}{\arg\!\max}
\usepackage{amsmath,amssymb,amsthm}
\usepackage{dsfont}

\newtheorem{theorem}{Theorem}
\newtheorem{proposition}{Proposition}

\newtheorem{assumption}{Assumption}
\newtheorem{lemma}{Lemma}
\title{Online MDP with Transition Prototypes: A Robust Adaptive Approach}
\author{
    Shuo Sun\textsuperscript{\rm 1},
    Meng Qi\textsuperscript{\rm 2},
    Zuo-Jun Max Shen\textsuperscript{\rm 1,3}
}
\affiliations{
    \textsuperscript{\rm 1}Department of Industrial Engineering and Operations Research, UC Berkeley, Berkeley, California 94720, USA\\
    \textsuperscript{\rm 2}SC Johnson College of Business, Cornell University, Ithaca, New York 14853, USA\\
    \textsuperscript{\rm 3}Faculty of Engineering and Faculty of Business and Economics, The University of Hong Kong, Hong Kong, China\\
    shuo\_sun@berkeley.edu, mq56@cornell.edu, maxshen@berkeley.edu
}
\usepackage{bibentry}
\begin{document}

\maketitle

\begin{abstract}
In this work, we consider an online robust Markov Decision Process (MDP) where we have the information of finitely many prototypes of the underlying transition kernel. We consider an adaptively updated ambiguity set of the prototypes and propose an algorithm that efficiently identifies the true underlying transition kernel while guaranteeing the performance of the corresponding robust policy. 
To be more specific, we provide a sublinear regret of the subsequent optimal robust policy. We also provide an early stopping mechanism and a worst-case performance bound of the value function. In numerical experiments, we demonstrate that our method outperforms existing approaches, particularly in the early stage with limited data. This work contributes to robust MDPs by considering possible prior information about the underlying transition probability and online learning, offering both theoretical insights and practical algorithms for improved decision-making under uncertainty.
\end{abstract}


\section{Introduction}\label{intro}
% Multi-model (and one short paragraph of application)\\
% then
% Multi-model is the transition probability, we consider a MDP setting. 

% Data is streaming, we consider the online setting\\
% then
% Robustness is essential to get the performance bound for value function\\
% then 
% Talk about our setting. Differentiate from existing setting and justify why our setting (regret, etc) makes more sense. (either here or later...comment on the technical novelty on discrete ambiguity set and early stopping)\\
% then
% Summarize result theoretical (comment on technical contribution) + mention extension to non-robust setting and the technical difference + numerical observation\\
% \sss{refer to multi-model paper}

Markov Decision Processes (MDPs) have become a fundamental framework for sequential decision-making under uncertainty, with applications spanning diverse fields such as control, healthcare and supply chain management. Despite their widespread use, MDPs often face challenges when the true transition dynamics are unknown, potentially leading to suboptimal decisions. 

In many real-world scenarios, decision-makers may rely on external datasets to parameterize the MDP model, but have access to multiple plausible model estimates, each potentially leading to different optimal policies. This setting is commonly seen in many applications, for example, the healthcare system \citep{steimle2021multi}. Consider the context of optimizing its breast cancer screening protocol. Decision-makers might have access to local hospital data, a national cancer research institute's model, and an international meta-analysis. Each source could suggest a different optimal screening frequency and age range for mammograms. This situation exemplifies the challenge of determining which model to trust or how to integrate insights from multiple sources to create a robust and effective policy when faced with various plausible model estimates. Similar challenges with multiple transition models arise in recommendation systems, supply chain management, and other domains where early performance and worst-case guarantees are crucial \citep{chatterjee2020multiple}. Moreover, the concept of multiple parameter models is analogous to the scenario-based stochastic programming literature, where each scenario represents a different possibility of the uncertain parameters. 

In this work, we focus on this multi-model setting where there are multiple models (prototypes) of the transition probabilities of the underlying Markov chain and the goal is to identify the true model and therefore solve for the optimal policy. Moreover, we address the problem in an online setting that we need to make real-time decisions with streaming data while knowing the prototypes. These prototypes could be estimated from offline dataset. The key challenge in such settings is two-fold: First, we need to efficiently identify the true underlying transition model while making decisions in real-time. Second, and perhaps more critically, we must ensure good performance during the learning phase when data is limited and model uncertainty is high. Classical online MDP algorithms focus primarily on achieving sublinear regret but may perform poorly in early stages and lack worst-case performance guarantees. 

% In this work, we aim to answer the following critical questions: (i) \textit{How can we identify the true choice model efficiently?} (ii) \textit{Can we mitigate the loss or potential harm due to model misspecification before identifying the true kernel?} (iii) \textit{How should we balance the exploration and exploitation in this setting?}

To address these challenges, we propose a novel robust learning algorithm that efficiently identifies the true transition kernel while guaranteeing model performance during the exploration stage. Our approach gradually updates the discrete prototype set and calculates the optimal robust policy, which achieves sublinear regret and provides a lower bound for the algorithm performance at each episode. As data accumulates, we propose a termination mechanism that efficiently identifies the true transition kernel.

Our work differentiates itself from existing approaches in several key aspects. First, we consider an online MDP with structural information of prototypes, which has not been studied before. Moreover, most work in robust MDP considers an offline setting or assumes access to a generator but we consider an online setting. 
Typically, robust MDP approaches assume a fixed ambiguity set size to calculate the optimal policy in the worst-case scenario. In contrast, we aim to optimize performance under the true model and gradually shrink the ambiguity set as data accumulates. This fundamental difference in goals sets us apart from existing methods that consider exogenous robustness, where the environment may be perturbed and the goal is to optimize for the worst-case scenario. In those works, the size of the uncertainty set is known, but the nominal transition probability is unknown. We, however, assume the existence of a true nominal system and design an adaptive robust algorithm that remains robust when data points are limited – what we term \textit{endogenous} robustness. Our ambiguity set shrinks as we collect more data.
Our work is closest to the online robust MDP work by \cite{dong2022online}. However, our work has an essential difference: they consider exogenous robustness and fix the size of the ambiguity set, whereas we aim to optimize the model performance under the true kernel.
It is important to emphasize that there are no existing sublinear regret results for online robust MDP problems, and achieving such results is notoriously difficult in general. In this work, by leveraging known prototypes of the underlying transition probability, we are able to provide sublinear regret bounds. This demonstrates the significant benefit of incorporating useful prior information about the underlying MDP model. Our approach could offer valuable insights for future work on model-based MDPs, particularly in scenarios where structural information is available or can be inferred.
The main contributions of our work are as follows:
\begin{enumerate}
    \item We propose a novel algorithm for learning robust policies in MDPs with multiple transition dynamic prototypes in an online setting (RPO-AAS). We show that our algorithm achieves sublinear regret with respect to the optimal policy for the true model and introduce an early stopping mechanism that allows our algorithm to converge to the true model more quickly with sufficient evidence.
    \item  We also propose a non-robust algorithm (NRPO-NPC) and analyze the technical performance guarantees. This algorithm does not calculate the robust optimal policy, but selects the prototype that is closest to the empirical distribution and runs the optimal policy corresponding to this prototype. Interestingly, we show that introducing robustness in the algorithm does not sacrifice efficiency.
\item Through numerical experiments, we demonstrate the effectiveness of our approach compared to existing methods, showing improved performance particularly in the early stage with limited data. 
\end{enumerate}

\section{Related Work}
Recent research has explored MDPs with parameter ambiguity using multiple models. \cite{steimle2021multi} and \cite{buchholz2019computation} consider finding a policy that maximizes a weighted performance across multiple models of MDPs. They proved NP-hardness of the problem and developed exact and approximate solution methods. \cite{ahmed2017sampling} explore sampling rewards and transition probabilities to generate a finite set of MDPs and find a policy to minimize the maximum regret over the set of MDPs. Our work differs from these approaches in two key aspects. Firstly, we consider an online setting, whereas previous works focused on offline setting. Secondly, our goal is to identify the true model and optimize its performance during exploration while guaranteeing robustness, rather than optimizing weighted performance for given weights or worst-case regret across all models.

The problem of regret minimization in MDPs with a fixed reward function has been studied extensively since \cite{burnetas1997optimal} and \cite{auer2006logarithmic}. Provably efficient learning algorithms fall into two main categories: The first applies \textit{optimism in the face of uncertainty} principle \citep{kearns2002near, brafman2002r, azar2017minimax} while the second utilizes posterior sampling reinforcement learning \citep{osband2013more, osband2017posterior}. \cite{agrawal2017posterior} combine these approaches, leveraging both the optimistic principle and posterior sampling to achieve a regret bound for weakly communicating MDPs. Currently the best regret bound for finite MDP is $\tilde{O}(\sqrt{H|\St||\A|T}+H^2 |\St|^2|\A|+H\sqrt{T})$ from the UCBVI algorithm, where $\St$ is the finite space of states, $\A$ is the set of finite actions and $H$ is the number of horizons \citep{azar2017minimax}. Despite these advancements, to our knowledge no existing work considers robust algorithms in MDPs with sublinear regret compared to the optimal reward.

%robusrt MDP
Robust MDPs consider the transition kernels that take values from an uncertainty set and learn an optimal robust policy that maximizes the worst-case value function. Most work in Robust MDP assumes that the the uncertainty set is known \cite{iyengar_robust_2005,nilim_robust_2005, xu_distributionally_2010}. Recently some work consider the robust optimal policy when the uncertainty set is not exactly known, or say reinforcement learning. Some work assumes that there is a generative model \citep{panaganti2022sample, yang2022toward} or assumes an offline dataset is present \citep{zhou2021finite,qi2020robust,kallus2022doubly,ma2022distributionally}. To our knowledge, only \cite{dong2022online} considers the robust policy learning in online setting. They propose algorithms that achieve a regret of $\tilde{O}(|\St||\A|^2H^2)$ under s-rectangular uncertainty set. 
However, these work have a different goal from our work. As discussed before, they consider the \textit{exogeneous} robustness, while we consider \textit{endogeneous} robustness. They consider an ambiguity set with fixed size while the radius of our algorithm is shrinking we when collect more data.

Another line of research characterizes the uncertainty through adversarial MDP formulations, where the environment parameters can be adversarially chosen. Most studies focus on the setting where only the reward function can be corrupted, while transition dynamics of the MDP remain fixed but potentially unknown \citep{neu2010online,cai2020provably, jin2020learning, rosenberg2019online,jin2020simultaneously,cai2020provably}. \cite{neu2010online} first proposes the online loop-free setting and show a regret of $\tilde{O}(L^2\sqrt{T|\A|}/\alpha)$ under some assumptions, where $L$ is the length of the longest path in the graph, $T$ is the number of episodes, and $\alpha$ is a probability parameter in the assumption. 
 Some work investigates settings where adversaries can corrupt transition metrics.  \citep{lykouris2021corruption} consider the setting that the transition is only allowed to be adversarially chosen for $C$ out of the $T$ total episodes and establish a regret of $\tilde{O}(C^2+\sqrt{T})$. Our prototype elimination approach shares similarities with arm-elimination methods in multi-armed bandit problems \citep{even2006action,audibert2010best}, but handles the additional complexity of state transitions rather than simple rewards.

\section{Problem Formulation and Preliminaries}
\subsection{Problem Formulation}
We consider a Markov Decision Process defined by a tuple $(\St, \A, P_0, r)$, where $\St$ is the finite state space and $\A$ is the finite action space, $P_0: \St \times \A \times \St \rightarrow [0,1]$ is the transition kernel, $r:\St\times \A\rightarrow \R$ is the reward function. More specifically, we use $P_0(s,a)$ and $r(s,a)$ to denote the probability distribution of the next state and immediate reward when taking action $a$ at state $s$. Let $P_0(s'|s,a)$ denote the probability of arriving at state $s'$ when choosing action $a$ at state $s$. Moreover, we assume the reward $r(s,a)$ is deterministic, and without loss of generality, $r(s,a)$  belongs to $[0,1]$. However, we would like to comment here that extending the algorithms to the setting with unknown reward does not add significant difficulty.

\paragraph{Loop-Free MDP.} In this work, we consider an episodic MDP with finite horizons. We assume the MDP has a \textit{loop-free} structure: The state space can be decomposed into $L+1$ non-intersecting layers $\St_0, \dots, \St_L$ such that $\St=\cup_{l=0}^L \St_l$, $\St_i\cap \St_j=\emptyset$ for $i\not= j$. Moreover, the first and the last layers are singletons, i.e., $\St_0=\{s_0\}$, $\St_L=\{s_L\}$. Let $\Layer(s)$ denote the layer of state $s$. The loop-free structure means the transitions are only possible between consecutive layers. These assumptions are not necessary, but are commonly adopted in literature, intended to simplify notation and analysis, and can be modified for a more general setup \citep{rosenberg2019online, jin2020learning}.
% \sss{We should be able to easily generalize our setting to the finite-horizon setting with changing probabilities and each time step has  prototype. Maybe reframe our work in this way.} 

\paragraph{Transition Prototypes.} 
In this work, we aim to illustrate the benefit of utilizing prior information about the transition probabilities. Specifically, we consider prototypes that are known to the decision-maker, each of which may correspond to an underlying model or mechanism that is driving the transition of the states.
We assume that for each layer $l$, there are $K_l$ prototypes of the transition kernel in the candidate set, denoted as $\{1,2,\dots, K_l\}$ and collectively referred to as $\K_l$. For any layer $l$, the transition probability at state $s$ and action $a$ defined by prototype $k\in \K_l$ is $P^k(s,a)$. The true transition kernel of each layer $l$, denoted as \(k^*_l\), must be one of the prototypes, meaning that $\otimes_{s\in \St_l, a\in \A} P_0(s,a)=\otimes_{s\in \St_l, a\in \A} P^{k^*_l}(s,a) $.

In the algorithm, we will update the candidate set of prototypes gradually, and we let $\K_{l,t}$ denote the set of prototypes in episode $t$. We update the set by removing the prototypes that are unlikely to be true as we collect more data. For the prototypes, we make the following structural assumption, which essentially states that if the gap between some kernels at a particular state $s$ in the layer and action $a$ is small, then the difference at other states in this layer and actions cannot be too large. 
\begin{assumption}
\label{a-1}
For any layer $l = 0, \dots, L$, any state $s\in \St_l$, action $a\in \A$, and any prototype $k\in \K_l$, if for some constant $u \in \mathbb{R}$, the $l_1$-norm $\|P^k (s,a)- P^{0}(s,a)\|_1 \leq u$, then there exists a constant $\gamma \in \mathbb{R}$ such that $\|P^k (s',a')- P^{0}(s',a')\|_1 \leq \gamma u$ for any other $s'\in \St_l,a'\in \A$.
\end{assumption}
Assumption 1 reflects that states within the same layer often share similar transition patterns, which is common in practice.
 The constant $\gamma$ quantifies the variability of transition probability differences across state-action pairs, while $u$ represents the magnitude of these differences for a reference state-action pair. Importantly, our theoretical results depend solely on $\gamma$, not on the absolute differences captured by $u$. This formulation provides flexibility in accommodating various MDP structures while maintaining analytical tractability. While this assumption helps establish theoretical guarantees, our numerical experiments in Section \ref{sec:exp-random} show that the algorithm maintains good performance even with random prototypes where this assumption may not hold. 

% \sss{Justify the assumptions}
In this paper, we use $\|\cdot\|_1$ to denote the $l_1$-norm between two transition probability vector. For any two transition kernels at state $s$ and action $a$, $P_0(s,a)$ and $P_1(s,a)$, we define $\|P_0(s,a), P_1(s,a)\|_1= \sum_{s'\in \St} |P_0(s'|s,a) - P_1(s'|s,a)| $. 
% Notation, examples

In each episode $t$, let $\pi_t$ denote the policy, which is a mapping from the state space $\St$ to action space $\A$. Given the transition kernel $P_0$ and policy $\pi_t$, the expected reward in episode $t$ is: 
\[
\mathbb{E}[\sum_{l=0}^{L-1}r(s_l,\pi_t(s_l))|P_0,\pi_t ], 
\]
where $s_l$ is the state visited in layer $l$ and episode $t$ and $\pi_t(s_l)$ is the corresponding action. 
Then, the total expected reward of the learner over $T$ episodes is:
\[R((\pi_t)_{t\in [T]},P_0)=\sum_{t=1}^T \mathbb{E}[\sum_{l=0}^{L-1}r(s_l,\pi_t(s_l))|P_0,\pi_t ].
\]

For a stationary policy $\pi$, with a slight abuse of notation, the total expected reward is given by
\[
R(\pi, P_0)=\sum_{t=1}^T \mathbb{E}[\sum_{l=0}^{L-1}r(s_l,\pi(s_l))|P_0,\pi ].
\]

Therefore, the regret can be defined as 
\[Reg = R(\pi^*,P_0)- R((\pi_t)_{t\in [T]} ,P_0).
\]
where $\pi^* \in \arg \max_{\pi} \mathbb{E}[\sum_{l=0}^{L-1}r(s_l,\pi(s_l))]$ is the optimal policy. 
% Prelims: occupancy  measure

Our regret definition diverges from that in the robust MDP literature \citep{dong2022online, zhou2021finite} which optimizes worst-case reward over an ambiguity set, with regret measured as the gap between worst-case rewards of the algorithm's policy and the optimal worst-case robust policy. In contrast, we optimize reward under the true transition kernel, aligning with the online MDP framework \citep{neu2010online}. 
\subsection{Preliminaries}

\paragraph{Occupancy measures.} 
We now reformulate the learner's problem using the concept of occupancy measures. We introduce occupancy measures for the purpose of analysis, which has been widely used in the analysis for loop-free MDP \citep{jin2020learning, rosenberg2019online}. Given a policy $\pi$ and transition kernel $P$, for any state $s\in \St_l$, $s'\in \St_{l+1}$, the occupancy measure $q^{P,\pi}$ is defined as: 
\[
q^{P,\pi}(s,a,s')=\prb[s_{\Layer(s)}=s, \pi(s)=a, s_{\Layer(s)+1}=s'|P,\pi].
\]

An occupancy measure satisfies the following two properties and these two properties suffice to define any function $q: \St \times \A \times \St \rightarrow [0,1]$ to be an occupancy measure. 1. The learner traverses every layer in each episode due to the loop-free structure, i.e., for every $l=0,\dots, L-1$, 
\[
\sum_{s\in \St_l} \sum_{a\in \A}\sum_{s'\in \St_{l+1}}q(s,a,s')=1.
\]
2. The probability of entering a state from the previous layer equals the probability of leaving it. 
% Due to the loop-free structure, the learner will reach a state in the next layer.
Thus, for every $l=1,\dots, L-1$ and $s\in \St_l$,
\[
\sum_{s'\in \St_{l+1}}\sum_{a\in \A}q(s,a,s')=\sum_{s'\in \St_{l-1}}\sum_{a\in \A}q(s',a,s).
\]

Given an occupancy measure $q$, the transition function $P^q$ and the policy $\pi^q$ can be induced as follows:
\[
P^q(s'|s,a)=\frac{q(s,a,s')}{\sum_{y\in \St_{\Layer(s)+1}}q(s,a,y) },
\]
\[
\pi^q(a|s)=\frac{\sum_{s'\in \St_{\Layer(s)+1}}q(s,a,s')}{\sum_{b\in \A}\sum_{s'\in \St_{\Layer(s)+1}}q(s,b,s')}.
\]
% From \cite{rosenberg2019online}, $P^{q^{P,\pi}}=P$ and $\pi^{q^{P,\pi}}=\pi$ for any $P$ and $\pi$. 
Then the problem of policy learning can be transformed to learning an occupancy measure $q_t\in \Delta(P_0)$ in each episode $t$, where $\Delta(P_0)$ is the set of all occupancy measures of an MDP with transition kernel $P_0$. With the definition of the occupancy measure, we redefine the expected reward and regret. The total expected reward of the learner is 
\begin{align*}
R((\pi_t)_{t\in [T]} ,P_0)&=\sum_{t=1}^T \mathbb{E}[\sum_{l=0}^{L-1}r(s_l,\pi_t(s_l))|P_0,\pi_t ]\\
&=\sum_{t=1}^T \langle q^{P_0,\pi_t},r \rangle
\end{align*}
Let $q^*\in \argmax_{q\in \Delta(P_0)} \sum_{t=1}^T\langle q^{P_0,\pi},r \rangle=q^{P_0,\pi^*}$ denote the occupancy measure corresponding to the optimal policy $\pi^*$ under $P_0$, the regret can be defined as 
\begin{align} \label{eq:reg}
Reg = & \max_{\pi} R(\pi,P_0)- R((\pi_t)_{t\in [T]} ,P_0) \\
= & \sum_{t=1}^T \langle q^*-q^{P_0,\pi_t}, r\rangle.
\end{align}

\section{The RPO-AAS Algorithm}
In this section, we introduce how we update the ambiguity set and calculate the robust optimal policy with respect to the ambiguity set in each episode. 
\begin{algorithm}
\caption{Robust Policy Optimization with Adaptive Ambiguity Set (RPO-AAS)}
\begin{algorithmic}[1] \label{alg:update-U}
\STATE Initialize: $\pi \leftarrow \pi_0$, number of samples $N_1(s,a)=0$ for each $s\in \St$, $a\in \A$
\FOR{$t = 1, \dots, T$}
    \FOR{$l = 1, \dots, L$}
        \STATE $s_{tl}, a_{tl} = \arg \max_{s \in \mathcal{S}_l, a \in \mathcal{A}} N_t(s,a)$
        \STATE Update the set of candidate prototypes:
        \STATE $\mathcal{K}_{l,t} = \{ k \in \mathcal{K}_{l,t-1} : \|P^k(s_{tl},a_{tl}) - \hat{P}_t(s_{tl},a_{tl})\|_1 \leq \sqrt{\frac{4|\mathcal{S}_{l+1}|\ln \frac{3LT}{\delta}}{N_{t}(s_{tl}, a_{tl})}} \}$
    \ENDFOR
    \STATE Update ambiguity set:
    \STATE $\mathcal{U}_t = \bigotimes_{s \in \mathcal{S}, a \in \mathcal{A}} \bigotimes_{k \in \mathcal{K}_{\mathcal{L}(s), t}} P^k(s,a)$
    \STATE Calculate optimal robust policy:
    \STATE $\pi_t =\arg \max_{\pi} \min_{P \in \mathcal{U}_t} R(\pi, P)$
    \STATE Execute policy $\pi_t$ for $L$ steps and obtain trajectory $s_l, a_l$ for $l = 1, \dots, L-1$
    \STATE $t = t+1$
    \STATE Update $N_t(s,a)$ for all $s,a$ and the empirical distribution $\hat{P}_t(s,a)$ for all $s, a$
\ENDFOR
\end{algorithmic}
\end{algorithm}
The algorithm initializes the policy $\pi$ to an arbitrary policy $\pi_0$ (e.g., a uniform policy) and sets the number of samples $N_1(s,a)$ to zero for each state-action pair $(s,a)$. In each episode, the following steps are performed:  First, for each layer $l = 1,\dots, L$, we identify the state-action pair $(s_{tl}, a_{tl})$ with the maximum number of samples in that layer. Next, we update the set of prototypes $\K_{l,t}$ by eliminating prototypes whose transition probabilities significantly deviate from the empirical transition distribution $\hat{P}_t(s,a)$ for the state-action pair $(s_{tl}, a_{tl})$. This update is crucial, as it relies on the state-action pair with the most occurrences, ensuring faster convergence of the empirical distribution to the true distribution. Subsequently, we update the ambiguity set $\mathcal{U}_t$ as the Cartesian product of the ambiguity sets for each state-action pair, where each set comprises the transition probabilities of the remaining prototypes in the corresponding layer. We then calculate the robust optimal policy $\pi_t$ by maximizing the worst-case value function over the ambiguity set $\mathcal{U}_t$. Since our ambiguity set satisfies the (s,a)-rectangular property, the optimal policy can be calculated using backward induction. The backward induction and ambiguity set update step takes $O(|S||A|+ \sum_{l=1}^L \mathcal{K}_l)$ time, which is efficient (details provided in the Appendix \ref{app:alg1}). Moreover, the key advantage of this ambiguity set construction is its high probability of including the true transition kernel as in the following lemma. 
\begin{lemma}
\label{lem:high-prob-P0} For the ambiguity set updated as described in Algorithm \ref{alg:update-U}, the true transition kernel lies in the ambiguity set  $\mathcal{U}_t$, i.e.,
$P_0\in  \mathcal{U}_t$ for all $t\in [T]$ with probability at least $1-\delta$.
\end{lemma}

We would like to point out that, this robust setting by considering the ambiguity set and solving for the worst-case value function over it allows one to have a worst-case performance bound, as stated in Proposition \ref{prop:1}. To be more specific, with the high-probability ambiguity set, we have that in each episode $t$, policy $\pi_t$ has the best worst-case performance and the performance of policy $\pi_t$ is lower bounded by the optimal objective value of the robust MDP. As we will see later, the non-robust algorithm lacks this robustness and could have poor performance, especially when we don't have enough data at the beginning.
\begin{proposition}
\label{prop:1}
In episode $t$, \(\min_{P \in \mathcal{U}_{t}} R(\pi_t, P)\ge \min_{P \in \mathcal{U}_{t}} R(\pi, P)\) for all policy $\pi$.
 Moreover, with probability at least \(1-\delta\), \(\min_{P \in \mathcal{U}_{t}} R(\pi_t, P)\) provides a lower bound for $R(\pi_t,P_0)$ with probability at least \(1-\delta\). 
\end{proposition}
The proof uses Hoeffding's inequality to bound the difference between the true and empirical transition probabilities. Due to space limitations, proofs for all results in this paper are provided in the appendix.
This proposition implies that in each episode $t$, policy $\pi_t$ has the best worst-case performance, and its actual performance is lower-bounded by the optimal objective value of the robust MDP. In contrast, a non-robust algorithm lacks this guarantee and may perform poorly, especially with limited data at the beginning. While the robust policy has its own advantages, the question remains that whether this robust policy has a good performance under the true transition kernel $P_0$. In the following section, we prove the theoretical guarantee of the RPO-AAS algorithm under $P_0$.

\section{Theoretical Results}
In this section, we first establish the sublinear regret bound, and then show the finite sample guarantee and the convergence result. 

\subsection{Analysis of Regret}
To bound the regret, we begin by decomposing \eqref{eq:reg} as follows:
\[Reg = 
\sum_{t=1}^T \langle q^*-q_t, r\rangle=\sum_{t=1}^T \langle q^*- \hat{q}_t, r\rangle  + \langle \hat{q}_t-q_t, r\rangle,
\]
where $q_t= q^{P_0,\pi_t}$, $\hat{q}_t=q^{P_t, \pi_t}$ and $\pi_t, P_t$ is the optimal solution of the robust optimization problem \(\max_{\pi}\min_{P \in \mathcal{U}_{t}} R(\pi, P)\). 
The high-level idea of our proof of regret involves three main steps. First, we upper bound the regret by the total reward difference between the true transition kernel $P_0$ and the kernel given by robust optimization $P_t$ under the optimal policy $\pi^*$ (Lemma \ref{lem:reg-dec}). We then bound this reward difference in two subsequent steps. We establish a bound on the one-norm difference between $P_0$ and $P_t$ (Lemma \ref{lem:bound-1-norm}), followed by bounding the difference in total reward given the bound of the difference in transition kernels (Lemma \ref{lem:7-1-err3}).

We begin with Lemma \ref{lem:reg-dec},  which provides an upper bound on the regret in terms of the total reward difference between the true transition kernel $P_0$ and the kernel given by robust optimization $P_t$ under the optimal policy $\pi^*$.
\begin{lemma}
\label{lem:reg-dec}
With probability at least $1-\delta$, $\sum_{t=1}^T \langle q^*- \hat{q}_t, r\rangle  + \langle \hat{q}_t-q_t, r\rangle\leq \sum_{t=1}^{T} \|q_t^{P_0,\pi^*}-q_t^{P_t,\pi^*}\|_1$.
\end{lemma}
% \begin{align*}
%   &  \langle q^*-q_t, r \rangle \\
%   & =  |\langle q^*, r-r_t \rangle+\langle q^*, r_t \rangle- \langle q_t, r_t \rangle+\langle q_t,r_t-r\rangle|\\
%     & \leq |\langle q^*-q_t, r-r_t \rangle| +|\langle q^*, r_t \rangle- \langle q_t, r_t \rangle| \\
%     & = |\langle q^*-q_t, r-r_t \rangle| +|\langle q^*-q_t, r_t \rangle| \\
% \end{align*}

Here, $\|q^{P_t,\pi^*}-q^{P_0,\pi^*}\|_1=\sum_{s,a,s'}|q^{P_t,\pi^*}(s,a,s')-q^{P_0,\pi^*}(s,a,s')|$. So it remains to bound $\sum_{t=1}^{T} \|q_t^{P_0,\pi^*}-q_t^{P_t,\pi^*}\|_1$. 
Based on the result from \cite{rosenberg2019online}, 
we bound \(\sum_{t=1}^{T} \|q^{P_0,\pi^*}-q^{P_t,\pi^*}\|_1 \) in the following lemma:  
\begin{lemma}\label{lem:bound-1-norm} For any policy $\pi$ and any $P_t\in \U_t$, with probability at least $1-\delta$, the following holds:
\begin{align*}&\sum_{t=1}^{T} \|q_t^{P_0,\pi}-q_t^{P_t,\pi}\|_1\\  &\leq 2\sum_{t=1}^T
\sum_{l=1}^L \sum_{m=0}^{l-1}\sum_{s_m \in \St_m}\sum_{a_m\in \A}q^{P_0,\pi}(s_m,a_m)\xi_t(s_m,a_m),
\end{align*}
 where $\xi_t(s,a)=\|P_t(\cdot|s,a),P_0(\cdot|s,a)\|_1$.
\end{lemma}

Thus, to bound the right-hand side in the lemma above, the key is to bound \(\xi_t(s,a)\), the difference in one-norm between $P_0$ and $P_t$.
\begin{lemma}
Suppose $P_0\in \U_t$. Then for any $s\in \St$, $a\in \A$, $t\in [T]$, and for all $k\in K_{t,\mathcal{L}(s)}$, we have:
\begin{align}
\label{eq:1norm-dif-2}
\|P_0(s,a),P^k(s,a) \|_1 \leq \sqrt{\frac{4|\St_{\mathcal{L}(s)+1}|  |\A| \ln \frac{3L T}{\delta}}{t}}
\end{align}
\label{lem:dif-p2}
\end{lemma}

With the established bound for $\xi_t$, we prove the following bound for the right-hand side of Lemma \ref{lem:bound-1-norm}. 

\begin{lemma}
\label{lem:7-1-err3} With probability at least $1-\delta$, the following holds:  
\begin{align*}
\sum_{t=1}^T\sum_{l=1}^L \sum_{m=0}^{l-1}\sum_{s_m \in \St_m}\sum_{a_m\in \A}q^{P_0,\pi}(s_m,a_m)\xi_t(s_m,a_m)\\  \leq L^2\gamma \sqrt{4T|\St|  |\A| \ln \frac{3L T}{\delta}}.
\end{align*}
\end{lemma}

By combining Lemma \ref{lem:reg-dec}, Lemma \ref{lem:bound-1-norm} and \ref{lem:7-1-err3}, we have the following regret bound:
\begin{theorem}
\label{thm:regret}
 With probability at least $1-\delta$, the RPO-AAS algorithm has the following regret bound:
\[
Reg\leq   L^2\gamma \sqrt{4T|\St|  |\A| \ln \frac{3L T}{\delta}}.
\]
\end{theorem}
It's worth noting that the state-of-the-art algorithm for general online MDPs achieves a regret bound of $\tilde{O}(\sqrt{H|\St| |\A|T}+H^2S^2|\A|+H\sqrt{T})$, where $H$ is the number of horizons \cite{azar2017minimax}. Our regret bound maintains the same dependence on $|\St|$, $|\A|$, and $T$. This demonstrates that, given structural information, our robust algorithm matches the efficiency of non-robust state-of-the-art approaches. However, it's important to note that designing efficient robust RL algorithms without structural information remains an open problem in the field.
% need to solve $T\ge \tau$, as $\tau$ is a constant dependent on $T$. 
% no we don't need to solve, see Theorem 1 in Shipra, we can include $T>f(T)$
\subsection{Finite-Sample Guarantee and Convergence}
In addition to the cumulative regret bound, we establish that the policy obtained by the proposed algorithm has a finite-sample performance guarantee and converges to the optimal policy. 

\begin{theorem}[Finite-sample guarantee] Let $v^{\pi}(s_0)$ denote the value function at state $s_0$ under policy $\pi$ under the true transition kernel. 
For any $\epsilon>0$, when $t\ge  \frac{4L^4 \gamma^2 |\St||\A| \ln \frac{3L T}{\delta}}{\epsilon^2}$, with probability at least $1-\delta$, $v^{\pi^*}(s_0)-v^{\pi_t}(s_0)\leq \epsilon$. 
\label{thm:finite}
\end{theorem}

This theorem states that after a sufficient number of episodes $t$, the value function of our algorithm's policy $\pi_t$ at the initial state $s_0$ is within $\epsilon$ of the optimal policy $\pi^*$'s value function, with high probability. The required number of episodes is inversely proportional to $\epsilon^2$. This dependency on $\epsilon^2$ is typical in many MDP problems \citep{panaganti2022sample}.

We next show that our algorithm can actually identify the true prototype after a finite number of episodes, leading to the optimal policy. 
\begin{theorem}[Prototype Ambiguity Set Convergence]
\label{thm:pro-convergence}
Let $h=\min_{s\in \St, a\in A, k\in [K]}\|P^k(\cdot|s,a),P_0(\cdot|s,a)\|_1$, then when $t\ge \frac{8|\St|^2|\A|\ln \frac{3LT}{\delta}}{h}$, the candidate set of prototypes only include the true prototypes, i.e., $K_{tl}=\{k^*_l\}$, thus $\pi_t=\pi^*$. 
\end{theorem}
This theorem establishes a finite-time guarantee for our algorithm's convergence to the true prototype and, consequently, the optimal policy. The result provides a principled stopping criterion, potentially improving the algorithm's practical efficiency.

\section{Extend to Non-robust Algorithm: Selecting the Best Candidate}
We propose another algorithm that selects the transition kernel that is nearest to the empirical distribution in each episode, referred to as non-robust policy optimization with nearest prototype-candidate(NRPO-NPC). Then in each episode, we run the optimal policy corresponding for the chosen transition kernel. We demonstrate that this approach provides the same theoretical performance guarantees for regret, convergence, and finite sample guarantees as the robust algorithm. However, it lacks the robustness guarantee.
To establish the theoretical results, we first decompose the regret at each episode as follows:
\begin{align}
\sum_{t=1}^T \langle q^*-q_t, r\rangle& = 
(q^{\pi^*,P_0}-q^{\pi^*, P_t})+(q^{\pi^*, P_t}-q^{\pi_t,P_t})\nonumber \\
&+(q^{\pi_t,P_t}-q^{\pi_t,P_0}) \label{eq:6}
\end{align}
The second term, $q^{\pi^*, P_t}-q^{\pi_t,P_t}\leq 0$, since $\pi_t$ is the optimal policy for transition kernel $P_t$. 
Similar to the proof for Theorem \ref{thm:regret}, we can bound the first term and the third term as long as we can bound the distance between $P_0$ and $P_t$, which is shown in the following lemma. 
\begin{lemma} 
For each layer $l$, let $s_{tl}, a_{tl} = \arg \max_{s \in \mathcal{S}_l, a \in \mathcal{A}} N_t(s,a)$ denote the (s,a) pair with the maximum number of samples in the layer. 
Let $k_t =\argmin_{ k\in K_{l,t}} \|P^k(s_{tl},a_{tl}) - \hat{P}_t(s_{tl},a_{tl})\|_1  $. Then for any $s\in \St$, $a\in \A$, $t\in [T]$, we have:
\begin{equation}
\label{eq:1norm-dif-2-nr}
\|P_0(\cdot |s,a),P^{k_t}(\cdot |s,a) \|_1\leq \sqrt{\frac{4|S_{\mathcal{L}(s)+1}|  |\A| \ln \frac{3L T}{\delta}}{t}}
\end{equation}
\label{lem:dif-p2-nr}
\end{lemma}
% The proof of this lemma and the other theoretical guarantees based on this lemma of are deferred to Appendix A.

\section{Numerical Experiments}
In the numerical experiments, we compare the performance of our proposed robust algorithm with the UCBVI algorithm \citep{azar2017minimax}, and the two benchmark algorithms that we propose that consider the prototype information. We will provide more details later. 

We consider a GridWorld experiment of size $5\times 4$, which is a widely used reinforcement setting from \cite{sutton1998reinforcement}. In each episode, the learner starts from the lower left corner and aims to the upper right corner. Let $(x_1,x_2)$ denote the coordinate, where $x_1\in \{0,1,2,3,4\}$ is the coordinate of the horizontal axis and $x_2\in \{0,1,2,3\}$ is the vertical axis coordinate. The learner collects rewards at some states, which we call reward states. We set the reward states to be $(2,2)$, $(1,1)$ and $(1,2)$ and the rewards are $3$, $5$ and $1$, respectively. At each state $s$ and $a$, the learner can either move up ($a= 0$) or right ($a=1$), with a success probability $z(s,a)$, and the learner goes to the opposite direction with probability $1- z(s,a)$. $z(s,a)$ is unknown. The learner's goal is to maximize the total collected rewards. If a learner reaches a boundary, she can only move inward. This problem is an episodic loop-free MDP, where each episode consists of $L=8$ layers. The number of states is $|\mathcal{S}|=20$ and the number of actions is $|\mathcal{A}|=2$.

\paragraph{Prototype configuration.} In each instance, we generate $K$ prototypes. We set $K=4$ and $K=10$, representing scenarios with few and many prototypes, respectively. For each prototype, we generate random $z_k(s,a)$ from a uniform distribution between $0$ and $1$. For simplicity, we generate different success probabilities only for different states, meaning $z_k(s,0)$ remains the same for all states, as does $z_k(s,1)$. We consider two types of prototype sets: The first set of prototype satisfies our assumption on the structure of transition prototypes (Assumption \ref{a-1}). Specifically, for any $s$ and $a$, we let $|z_{k_1}(s,a)-z_{k_2}(s, a)|$ be fixed for any kernel $k_1$ and $k_2$. We call this setting \textit{fixed-gap} prototypes. The second set does not satisfy this assumption. In this setting, we generate $z_k(s,0)$ and $z_k(s,1)$ for all prototypes randomly. We define this setting as \textit{random} prototypes.

\paragraph{Algorithms.} We compare four algorithms: (1) our robust algorithm (RPO-AAS), (2) UCBVI algorithm, (3) the non-robust nearest prototype-candidate algorithm (NRPO-NPC), and (4) its variant, NRPO-NPC-2. The latter is a heuristic that selects the prototype with the smallest 1-norm distance to the empirical transition probabilities across all states and actions in the layer. Details are provided in the Appendix \ref{app:alg_nr}.

\paragraph{Experiment Environment.} We conduct the numerical experiment using \textit{rlberry}, a Python library for reinforcement learning \citep{rlberry}. 
For each setting, we run 100 simulations. In each simulation, we record the average expected rewards in each episode. We then take the average of these simulations. The expected episode reward is the expectation of the total reward under the policy in episode $t$.

\subsection{Structured Prototypes Setting}
\begin{figure}[ht]
    \centering
    \includegraphics[width=0.8\linewidth]{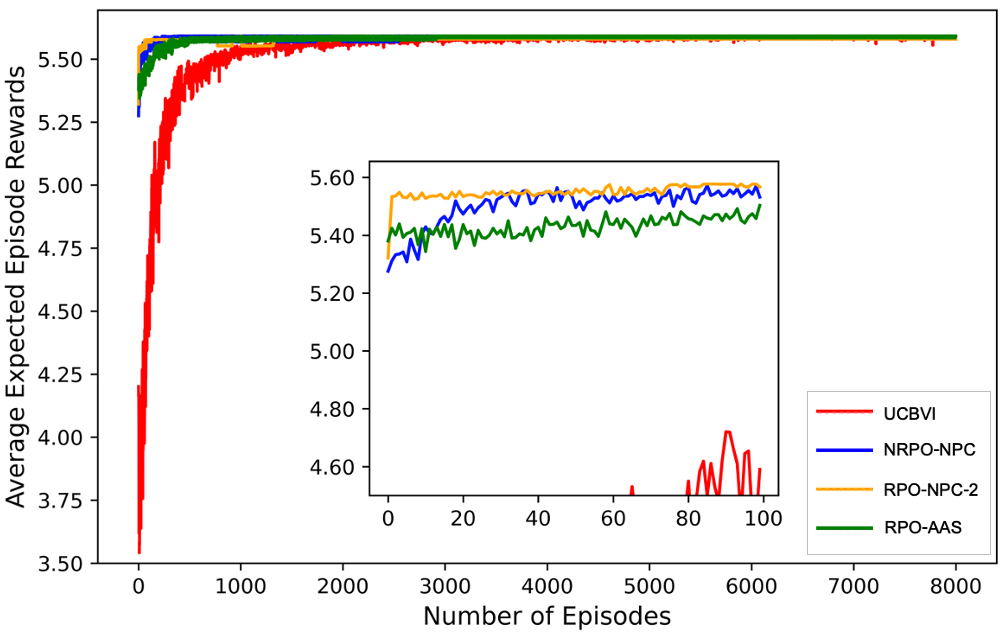}
    \caption{Average Expected Episode Rewards of different algorithms with Fixed-gap Prototypes when $K=4$.}
    \label{fig:struc-1}
\end{figure}
\begin{figure}[ht]
    \centering
    \includegraphics[width=0.8\linewidth]{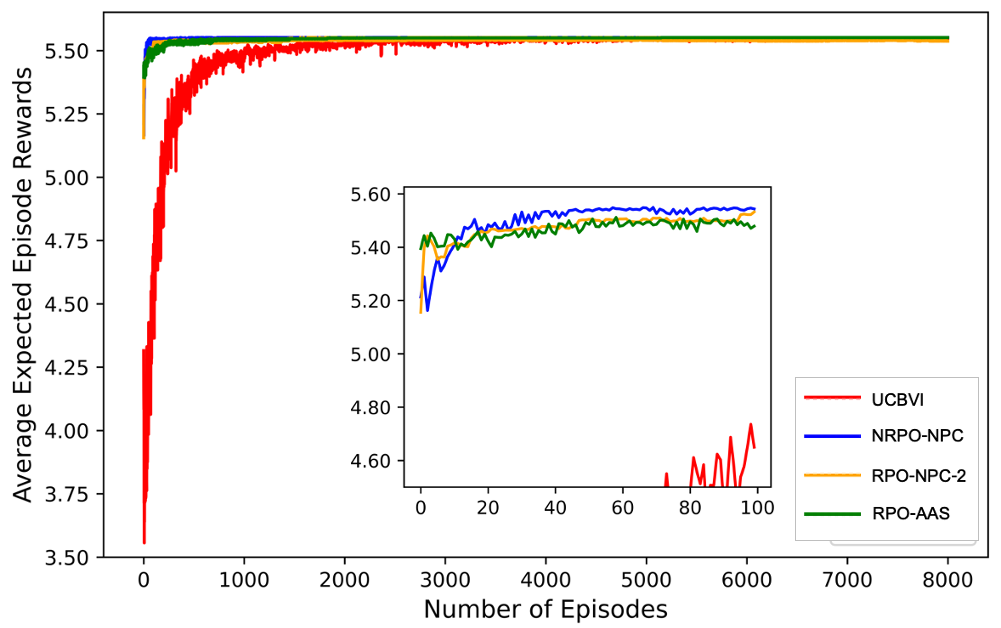}
    \caption{Average Expected Episode Rewards of different algorithms with Fixed-gap Prototypes when $K=10$.}
    \label{fig:struc-2}
\end{figure}

In the fixed-gap setting with $K=4$, we observe that NRPO-NPC, NRPO-NPC-2, and RO perform significantly faster than the UCBVI algorithm. This indicates that our proposed algorithms can leverage the prototype information effectively, resulting in better performance. NRPO-NPC-2 converges to the optimal policy fastest, although it lacks theoretical guarantees. When $K=10$, the performance of NRPO-NPC and RO surpasses that of NRPO-NPC-2 and UCBVI. Notably, in both cases, RO demonstrates better performance at the beginning, showcasing the advantage of considering robustness. 

Next, we compare algorithm performances when we relax the assumption and consider random prototypes. 
\subsection{Random Prototypes Setting} \label{sec:exp-random}
\begin{figure}[ht]
    \centering
    \includegraphics[width=0.8\linewidth]{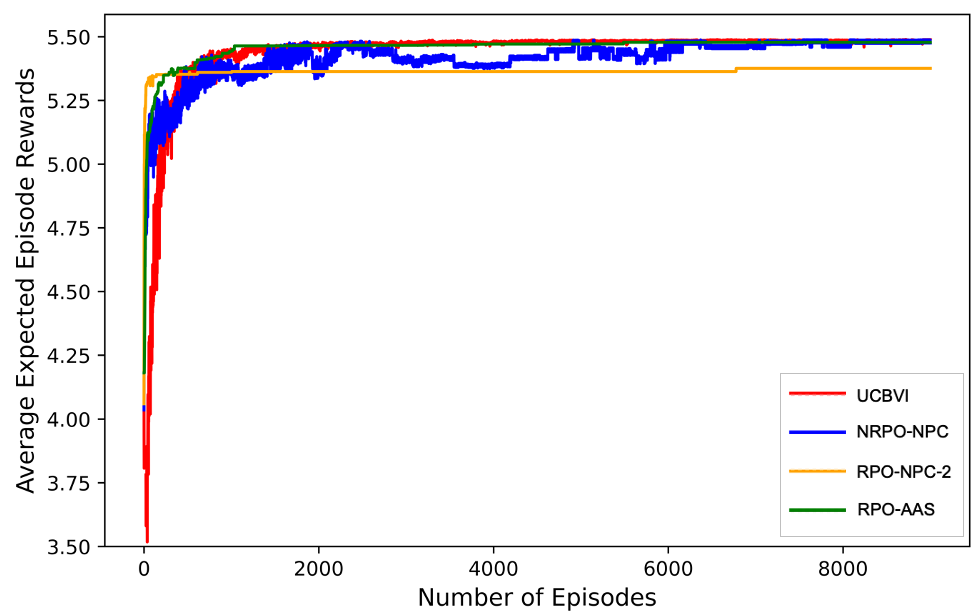}
    \caption{Average Expected Episode Rewards of different algorithms with 4 Random Prototypes.}
    \label{fig:random-exp}
\end{figure}

We start from $K=4$ prototypes. Figure \ref{fig:random-exp} shows the performance of the algorithms. In this setting, NRPO-NPC-2 couldn't converge to the optimal policy. RO yields better policies than UCBVI and NRPO-NPC in the first 2,000 episodes. Moreover, the policy given by RO has lower fluctuations than NRPO-NPC and UCBVI. NRPO-NPC outperforms UCBVI initially but shows greater variance and converges to the optimal solution more slowly than UCBVI.

When we increase the number of prototypes to 10, NRPO-NPC, NRPO-NPC-2, and RO continue to outperform UCBVI during the first 400 episodes. RO maintains the lowest variance, indicating that it yields the most stable policy. However, UCBVI converges to the optimal policy more rapidly than RO and NRPO-NPC in many cases, resulting in slightly superior performance after 2,000 episodes.

This observation suggests that as the number of prototypes increases, the benefits of incorporating prototype information diminish. This is logical, as in the limit of infinite prototypes, the algorithm would gain no advantage from prototype information. From Theorem \ref{thm:pro-convergence}, more prototypes potentially reduce $h$ and thus slow convergence, while too few prototypes may fail to include the transition kernel. Therefore, the number of prototypes $K$ presents a practical trade-off. Nevertheless, the RO algorithm maintains its robustness even in this many-prototype setting.

\begin{figure}[ht]
    \centering
    \includegraphics[width=0.8\linewidth]{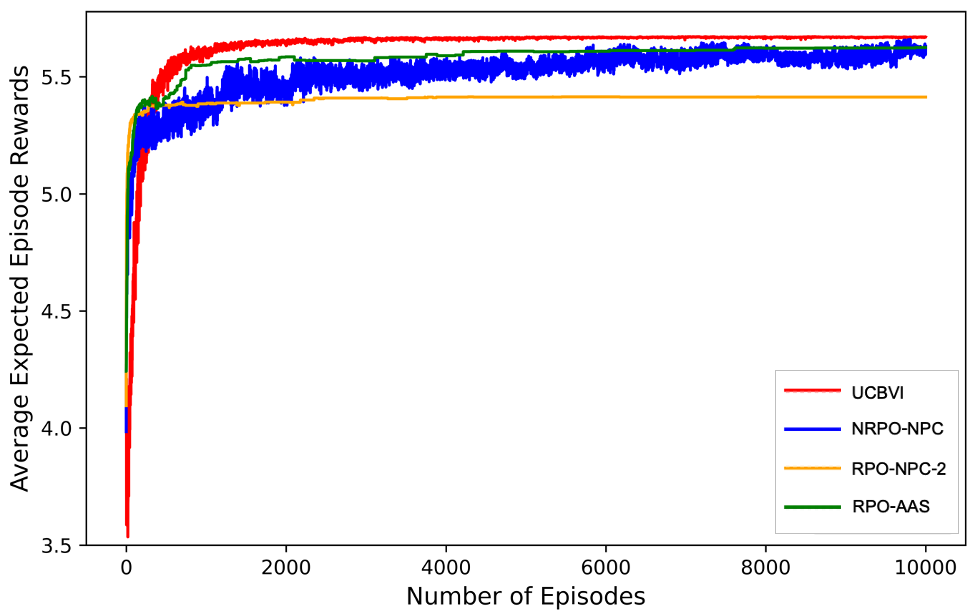}
    \caption{Average Expected Episode Rewards of different algorithms with 10 Random Prototypes.}
    \label{fig:enter-label}
\end{figure}

\section{Conclusion}
In this work, we introduced a novel approach for online MDPs with transition prototypes. Our robust adaptive algorithm efficiently identifies the true transition kernel while guaranteeing performance through robust policies. Theoretical analysis shows that the algorithm achieves sublinear regret, provides finite-sample guarantees, and converges to the optimal policy in finite time. Numerical experiments demonstrate its practical advantages, particularly in early learning stages and with structured prototypes. We also extended our analysis to a non-robust algorithm, highlighting the value of prototype information. This work shows the potential of the combination of structural information and robust optimization in reinforcement learning. Future work could explore extensions to more complex MDP settings and investigate robustness-optimality trade-offs in various applications.

\newpage
\bibliography{references}

\begin{thebibliography}{32}
\providecommand{\natexlab}[1]{#1}

\bibitem[{Agrawal and Jia(2017)}]{agrawal2017posterior}
Agrawal, S.; and Jia, R. 2017.
\newblock Posterior sampling for reinforcement learning: worst-case regret bounds.
\newblock \emph{arXiv preprint arXiv:1705.07041}.

\bibitem[{Ahmed et~al.(2017)Ahmed, Varakantham, Lowalekar, Adulyasak, and Jaillet}]{ahmed2017sampling}
Ahmed, A.; Varakantham, P.; Lowalekar, M.; Adulyasak, Y.; and Jaillet, P. 2017.
\newblock Sampling based approaches for minimizing regret in uncertain Markov decision processes (MDPs).
\newblock \emph{Journal of Artificial Intelligence Research}, 59: 229--264.

\bibitem[{Audibert and Bubeck(2010)}]{audibert2010best}
Audibert, J.-Y.; and Bubeck, S. 2010.
\newblock Best arm identification in multi-armed bandits.
\newblock In \emph{COLT-23th Conference on learning theory-2010}, 13--p.

\bibitem[{Auer and Ortner(2006)}]{auer2006logarithmic}
Auer, P.; and Ortner, R. 2006.
\newblock Logarithmic online regret bounds for undiscounted reinforcement learning.
\newblock \emph{Advances in neural information processing systems}, 19.

\bibitem[{Azar, Osband, and Munos(2017)}]{azar2017minimax}
Azar, M.~G.; Osband, I.; and Munos, R. 2017.
\newblock Minimax regret bounds for reinforcement learning.
\newblock In \emph{International conference on machine learning}, 263--272. PMLR.

\bibitem[{Brafman and Tennenholtz(2002)}]{brafman2002r}
Brafman, R.~I.; and Tennenholtz, M. 2002.
\newblock R-max-a general polynomial time algorithm for near-optimal reinforcement learning.
\newblock \emph{Journal of Machine Learning Research}, 3(Oct): 213--231.

\bibitem[{Buchholz and Scheftelowitsch(2019)}]{buchholz2019computation}
Buchholz, P.; and Scheftelowitsch, D. 2019.
\newblock Computation of weighted sums of rewards for concurrent MDPs.
\newblock \emph{Mathematical Methods of Operations Research}, 89: 1--42.

\bibitem[{Burnetas and Katehakis(1997)}]{burnetas1997optimal}
Burnetas, A.~N.; and Katehakis, M.~N. 1997.
\newblock Optimal adaptive policies for Markov decision processes.
\newblock \emph{Mathematics of Operations Research}, 22(1): 222--255.

\bibitem[{Cai et~al.(2020)Cai, Yang, Jin, and Wang}]{cai2020provably}
Cai, Q.; Yang, Z.; Jin, C.; and Wang, Z. 2020.
\newblock Provably efficient exploration in policy optimization.
\newblock In \emph{International Conference on Machine Learning}, 1283--1294. PMLR.

\bibitem[{Chatterjee et~al.(2020)Chatterjee, Chmel{\'\i}k, Karkhanis, Novotn{\`y}, and Royer}]{chatterjee2020multiple}
Chatterjee, K.; Chmel{\'\i}k, M.; Karkhanis, D.; Novotn{\`y}, P.; and Royer, A. 2020.
\newblock Multiple-environment markov decision processes: Efficient analysis and applications.
\newblock In \emph{Proceedings of the International Conference on Automated Planning and Scheduling}, volume~30, 48--56.

\bibitem[{Domingues et~al.(2021)Domingues, Flet-Berliac, Leurent, M{\'e}nard, Shang, and Valko}]{rlberry}
Domingues, O.~D.; Flet-Berliac, Y.; Leurent, E.; M{\'e}nard, P.; Shang, X.; and Valko, M. 2021.
\newblock {rlberry - A Reinforcement Learning Library for Research and Education}.

\bibitem[{Dong et~al.(2022)Dong, Li, Wang, and Zhang}]{dong2022online}
Dong, J.; Li, J.; Wang, B.; and Zhang, J. 2022.
\newblock Online policy optimization for robust MDP.
\newblock \emph{arXiv preprint arXiv:2209.13841}.

\bibitem[{Even-Dar et~al.(2006)Even-Dar, Mannor, Mansour, and Mahadevan}]{even2006action}
Even-Dar, E.; Mannor, S.; Mansour, Y.; and Mahadevan, S. 2006.
\newblock Action elimination and stopping conditions for the multi-armed bandit and reinforcement learning problems.
\newblock \emph{Journal of machine learning research}, 7(6).

\bibitem[{Iyengar(2005)}]{iyengar_robust_2005}
Iyengar, G.~N. 2005.
\newblock Robust {Dynamic} {Programming}.
\newblock \emph{Mathematics of Operations Research}, 30(2): 257--280.

\bibitem[{Jin et~al.(2020)Jin, Jin, Luo, Sra, and Yu}]{jin2020learning}
Jin, C.; Jin, T.; Luo, H.; Sra, S.; and Yu, T. 2020.
\newblock Learning adversarial markov decision processes with bandit feedback and unknown transition.
\newblock In \emph{International Conference on Machine Learning}, 4860--4869. PMLR.

\bibitem[{Jin and Luo(2020)}]{jin2020simultaneously}
Jin, T.; and Luo, H. 2020.
\newblock Simultaneously learning stochastic and adversarial episodic mdps with known transition.
\newblock \emph{Advances in neural information processing systems}, 33: 16557--16566.

\bibitem[{Kallus et~al.(2022)Kallus, Mao, Wang, and Zhou}]{kallus2022doubly}
Kallus, N.; Mao, X.; Wang, K.; and Zhou, Z. 2022.
\newblock Doubly robust distributionally robust off-policy evaluation and learning.
\newblock In \emph{International Conference on Machine Learning}, 10598--10632. PMLR.

\bibitem[{Kearns and Singh(2002)}]{kearns2002near}
Kearns, M.; and Singh, S. 2002.
\newblock Near-optimal reinforcement learning in polynomial time.
\newblock \emph{Machine learning}, 49: 209--232.

\bibitem[{Lykouris et~al.(2021)Lykouris, Simchowitz, Slivkins, and Sun}]{lykouris2021corruption}
Lykouris, T.; Simchowitz, M.; Slivkins, A.; and Sun, W. 2021.
\newblock Corruption-robust exploration in episodic reinforcement learning.
\newblock In \emph{Conference on Learning Theory}, 3242--3245. PMLR.

\bibitem[{Ma et~al.(2022)Ma, Liang, Blanchet, Liu, Xia, Zhang, Zhao, and Zhou}]{ma2022distributionally}
Ma, X.; Liang, Z.; Blanchet, J.; Liu, M.; Xia, L.; Zhang, J.; Zhao, Q.; and Zhou, Z. 2022.
\newblock Distributionally robust offline reinforcement learning with linear function approximation.
\newblock \emph{arXiv preprint arXiv:2209.06620}.

\bibitem[{Neu et~al.(2010)Neu, Gy{\"o}rgy, Szepesv{\'a}ri et~al.}]{neu2010online}
Neu, G.; Gy{\"o}rgy, A.; Szepesv{\'a}ri, C.; et~al. 2010.
\newblock The Online Loop-free Stochastic Shortest-Path Problem.
\newblock In \emph{COLT}, volume 2010, 231--243. Citeseer.

\bibitem[{Nilim and El~Ghaoui(2005)}]{nilim_robust_2005}
Nilim, A.; and El~Ghaoui, L. 2005.
\newblock Robust {Control} of {Markov} {Decision} {Processes} with {Uncertain} {Transition} {Matrices}.
\newblock \emph{Operations Research}, 53(5): 780--798.

\bibitem[{Osband, Russo, and Van~Roy(2013)}]{osband2013more}
Osband, I.; Russo, D.; and Van~Roy, B. 2013.
\newblock (More) efficient reinforcement learning via posterior sampling.
\newblock \emph{Advances in Neural Information Processing Systems}, 26.

\bibitem[{Osband and Van~Roy(2017)}]{osband2017posterior}
Osband, I.; and Van~Roy, B. 2017.
\newblock Why is posterior sampling better than optimism for reinforcement learning?
\newblock In \emph{International conference on machine learning}, 2701--2710. PMLR.

\bibitem[{Panaganti and Kalathil(2022)}]{panaganti2022sample}
Panaganti, K.; and Kalathil, D. 2022.
\newblock Sample complexity of robust reinforcement learning with a generative model.
\newblock In \emph{International Conference on Artificial Intelligence and Statistics}, 9582--9602. PMLR.

\bibitem[{Qi and Liao(2020)}]{qi2020robust}
Qi, Z.; and Liao, P. 2020.
\newblock Robust batch policy learning in markov decision processes.
\newblock \emph{arXiv preprint arXiv:2011.04185}.

\bibitem[{Rosenberg and Mansour(2019)}]{rosenberg2019online}
Rosenberg, A.; and Mansour, Y. 2019.
\newblock Online convex optimization in adversarial markov decision processes.
\newblock In \emph{International Conference on Machine Learning}, 5478--5486. PMLR.

\bibitem[{Steimle, Kaufman, and Denton(2021)}]{steimle2021multi}
Steimle, L.~N.; Kaufman, D.~L.; and Denton, B.~T. 2021.
\newblock Multi-model Markov decision processes.
\newblock \emph{IISE Transactions}, 1--16.

\bibitem[{Sutton and Barto(1998)}]{sutton1998reinforcement}
Sutton, R.~S.; and Barto, A.~G. 1998.
\newblock Reinforcement learning: an introduction MIT Press.
\newblock \emph{Cambridge, MA}, 22447: 10.

\bibitem[{Xu and Mannor(2010)}]{xu_distributionally_2010}
Xu, H.; and Mannor, S. 2010.
\newblock Distributionally {Robust} {Markov} {Decision} {Processes}.
\newblock In Lafferty, J.~D.; Williams, C. K.~I.; Shawe-Taylor, J.; Zemel, R.~S.; and Culotta, A., eds., \emph{Advances in {Neural} {Information} {Processing} {Systems} 23}, 2505--2513. Curran Associates, Inc.

\bibitem[{Yang, Zhang, and Zhang(2022)}]{yang2022toward}
Yang, W.; Zhang, L.; and Zhang, Z. 2022.
\newblock Toward theoretical understandings of robust markov decision processes: Sample complexity and asymptotics.
\newblock \emph{The Annals of Statistics}, 50(6): 3223--3248.

\bibitem[{Zhou et~al.(2021)Zhou, Zhou, Bai, Qiu, Blanchet, and Glynn}]{zhou2021finite}
Zhou, Z.; Zhou, Z.; Bai, Q.; Qiu, L.; Blanchet, J.; and Glynn, P. 2021.
\newblock Finite-sample regret bound for distributionally robust offline tabular reinforcement learning.
\newblock In \emph{International Conference on Artificial Intelligence and Statistics}, 3331--3339. PMLR.

\end{thebibliography}

\newpage
\appendix
\section{Additional Analysis of Algorithm 1}\label{app:alg1}\paragraph{Backward Induction.}
At each episode $t$, we calculate the robust policy $\pi_t$ by maximizing the worst-case value function over our ambiguity set $\mathcal{U}_t$ through backward induction. Starting from layer $L-1$, for each state $s$ and action $a$, we compute:
\[
Q(s,a) = \min_{P \in \mathcal{U}_t} [r(s,a) + \sum_{s'} P(s'|s,a)V(s')]
\]
\[
V(s) = \max_{a} Q(s,a)
\]
This process continues backward through all layers to the initial state. The computational complexity is $O(|\mathcal{S}||\mathcal{A}K) $ where $K$ is the maximum number of prototypes in any layer, since for each state-action pair we evaluate $K$ possible transitions
\citep{nilim_robust_2005,iyengar_robust_2005}
\paragraph{Computational Complexity.}
The computational complexity of RPO-AAS has three main components. First, the state-action pair identification requires $O(|\mathcal{S}||\mathcal{A})$ operations per layer. The ambiguity set update takes $O(\sum_{l=1}^L \mathcal{K}_l)$ operations since we update one pair per layer. Finally, the backward induction for policy computation requires $O(K|\mathcal{S}||\mathcal{A})$ operations due to the (s,a)-rectangular property.

\section{Additional Proofs}
\subsection{Proof of Lemma 1}
\begin{proof}[Proof of Lemma 1] By Hoeffding's inequality, we have that, the following inequality holds with probability at least $1-\delta$ for state $s_{tl}\in \St_l$, action $a_{tl}\in \A$ and each layer $l$: 
\begin{equation}
\label{eq:1norm-dif}\|P_0(\cdot |s,a),\hat{P}_t(\cdot |s,a) \|_1\leq \sqrt{\frac{4|\St_{\mathcal{L}(s)+1}|\ln \frac{3LT}{\delta}}{\min\{N_{t-1}(s,a),1\}}}
\end{equation}
For any layer $l$, recall that we define the set $\K_{l,t}$ as:
\begin{align*}&\K_{l,t} = \{k \in \K_{l,t-1} : \|P^k(s_{tl},a_{tl}), \hat{P}_{t}(s_{tl},a_{tl})\|_1 \\& \leq \sqrt{\frac{4|\St_{\mathcal{L}(s)+1}|\ln \frac{3LT}{\delta}}{N_{t-1}(s_{tl}, a_{tl})}} \}\end{align*}
This set includes all prototypes whose distance to the empirical transition kernel at $(s_{tl}, a_{tl})$ is at most $\sqrt{\frac{4|\St_{\mathcal{L}(s)+1}|\ln \frac{3LT}{\delta}}{N_{t-1}(s_{tl}, a_{tl})}}$.
Under the condition that inequality \eqref{eq:1norm-dif} holds, we know that $P_0(s_{tl},a_{tl})= P^k(s_{tl},a_{tl})$ for some $k\in \K_{l,t}$. Since the transition probabilities at the state and action in the same layer are derived from the same prototype, we have $P_0(s,a)\in \mathcal{U}_t(s,a)$ for all $s\in \St_l$ and $a\in \A$. Since inequality \eqref{eq:1norm-dif} holds with probability at least $1-\delta$, we conclude that $P_0 \in \mathcal{U}_t$ for all $t \in [T]$ with probability at least $1-\delta$.
\end{proof}

\subsection{Proof of Proposition 1}
\begin{proof}[Proof of Proposition 1]
From the definition of $\pi_t$,
     \(\min_{P \in \mathcal{U}_{t}} R(\pi_t, P) =\max_{\pi}\min_{P \in \mathcal{U}_{t}} R(\pi, P)\ge \min_{P \in \mathcal{U}_{t}} R(\pi, P)\) for any policy $\pi$. Moreover, $P_0\in \U_t$, then \( \min_{P \in \mathcal{U}_{t}} R(\pi_t, P) \leq R(\pi_t,P_0)\). Since $P_0\in \U_t$ with probability at least \(1-\delta\) (Lemma 1), we have that \(\max_{\pi}\min_{P \in \mathcal{U}_{t}} R(\pi, P)\) provides a lower bound for $R(\pi_t,P_0)$ with probability at least \(1-\delta\). 
\end{proof}

\subsection{Proof of Lemma 2}
\begin{proof}[Proof of Lemma 2]
By Lemma 1, we know that $P_0 \in \mathcal{U}_t$ with probability at least $1-\delta$. The subsequent analysis is conducted under the condition that $P_0 \in \mathcal{U}_t$, and therefore, the following results hold with probability at least $1-\delta$.

For the first term,
$$
\begin{aligned}\langle q^*- \hat{q}_t, r\rangle 
&=\max_{\pi}R(\pi, P_0)-\min_{P\in \mathcal{U}_t}R(\pi_t, P)  \\
&\leq R(\pi^*,P_0)-\min_{P\in \mathcal{U}_t}R(\pi^*, P) \\
&=  R(\pi^*,P_0)- R(\pi^*, P_t)  \\
&=\langle q^{P_0,\pi^*}-q^{P_t,\pi^*}, r\rangle.
\end{aligned}
$$

The inequality holds since $\pi^*$ is a feasible solution of $\max_{\pi}\min_{P\in \mathcal{U}_t}R(\pi, P)$. Thus, $\min_{P\in \mathcal{U}_t} R(\pi_t,P)= \max_{\pi}\min_{P\in \mathcal{U}_t}R(\pi, P) \geq \min_{P\in \mathcal{U}_t}R(\pi, P)$.

For the second term, we have $\langle \hat{q}_t-q_t, r\rangle = \min_{P\in \mathcal{U}_t}R(\pi_t, P) - R(\pi_t, P_0) \leq 0$ since $P_0\in \mathcal{U}_t$.
Therefore, we have 
\begin{align*}
\sum_{t=1}^T \langle q^*- \hat{q}_t, r\rangle  + \langle \hat{q}_t-q_t, r\rangle &\leq \sum_{t=1}^T \langle q^*- \hat{q}_t, r\rangle \\
&\leq \sum_{t=1}^T \langle q^{P_0,\pi^*}-q^{P_t,\pi^*}, r\rangle.
\end{align*}

Since $r(s,a)\in [0,1]$ for all $s\in \St$, $a\in \A$, it follows that \[
 \sum_{t=1}^T \langle q^{P_0,\pi^*} - q^{P_t,\pi^*}, r\rangle\leq \sum_{t=1}^{T} \|q^{P_0,\pi^*}-q^{P_t,\pi^*}\|_1. 
\]
\end{proof}

\subsection{Proof of of Lemma 3}
This proof is a straightforward combination of the following two lemmas in \cite{rosenberg2019online}. 
\begin{lemma}[Lemma B.1 in \cite{rosenberg2019online}]
For any policy $\pi$,
\begin{align}
\label{eq:a-lem-1}
\sum_{t=1}^T \|q^{P_t,\pi}-q^{P_0,\pi}\|_1 &\leq \sum_{t=1}^T \sum_{s\in \St}\sum_{a\in \A}
|q^{P_t,\pi}(s,a)-q^{P_0,\pi}(s,a)|\nonumber \\ &+\sum_{t=1}^T\sum_{s\in \St}\sum_{a\in \A}q^{P_0,\pi}(s,a)\xi_t(s,a) 
\end{align}
\label{lem:a-lem-1}
\end{lemma}
\begin{lemma}\label{lem:a-lem-2}[Lemma B.2 in \cite{rosenberg2019online}] For any policy $\pi$, and any $l=1, \dots, L-1$ and $t=1,\dots, T$, it holds that
\begin{align}
\label{eq:a-lem-2}
\sum_{s_l\in \St_l}\sum_{a_l\in \A} |q^{P_t,\pi}(s_l,a_l)-q^{P_0,\pi}(s_l,a_l) |\nonumber \\ \leq\sum_{m=0}^{l-1}\sum_{s_m \in \St_m}\sum_{a_m\in \A}q^{P_0,\pi}(s_m,a_m)\xi_t(s_m,a_m)
\end{align}
\end{lemma} 
Combining inequalities \eqref{lem:a-lem-1} and \eqref{lem:a-lem-2}, we prove Lemma 3.
\subsection{Proof of Lemma 4}
\begin{proof}[Proof of Lemma 4]
Let $l=\mathcal{L}(s)$, suppose $P_0\in \U_t$, from the definition of $\U_t$, for any $s\in \St_l$, $l \in [L]$, we have:
\[
\|P_0(\cdot |s,a),\hat{P}_t(\cdot |s,a) \|_1\leq \sqrt{\frac{4|\St_{l+1}|\ln \frac{3LT}{\delta}}{\min\{N_{t-1}(s,a),1\}}}
\]
For all $k\in K_{t,l}$, we bound the distance between $P_0$ and $P^k$ using triangle inequality:
\begin{align*}
&\|P_0(\cdot |s,a),P^k(\cdot |s,a) \|_1\\
&\leq\|\hat{P}_t(s,a),P^k(s,a)\|_1+ \|P_0(\cdot |s,a),\hat{P}_t(\cdot |s,a) )\|_1 \\ 
&\leq 4\sqrt{\frac{|\St_{l+1}|\ln \frac{3LT}{\delta}}{\min\{N_{t-1}(s,a),1\}}}
\end{align*}
Recall that $s_{tl}, a_{tl}$ is the state-action pair with the maximum number of samples for all the states in layer $l$. Since for any layer $l$, from the pigeonhole principle, we have \( N_{t-1}(s_{tl}, a_{tl})\ge \frac{t}{|\St_l||\A|}\). Thus, \[
\|P_0(s_{tl},a_{tl}),P^k(s_{tl},a_{tl}) \|_1\leq \sqrt{\frac{4|\St_{l+1}||\St_l||\A|\ln \frac{3LT}{\delta}}{t}}.
\]
From Assumption 1, for any $s\in \St_l$, $a\in \A$,
\begin{align*}
\|P_0(s,a),P^k(s,a)\|_1& \leq \gamma \|P_0(s_{tl},a_{tl}),P^k(s_{tl},a_{tl}) \|_1 \\
&\leq 4\gamma \sqrt{\frac{|\St_{l+1}||\St_l||\A|\ln \frac{3LT}{\delta}}{t}}.
\end{align*}
\end{proof}
\subsection{Proof of Lemma 5}
\begin{proof}[Proof of Lemma 5]
Due to the loop-free structure, each layer will be visited in each episode, thus for any layer $m$, $\sum_{s_m \in \St_m}\sum_{a_m\in \A}q^{P_0,\pi}(s_m,a_m)=1$. Combining this equation with Lemma 4, we obtain the following inequality: 
\begin{align*}
\label{eq:qp0pi}
& \sum_{s_m \in \St_m}\sum_{a_m\in \A}q^{P_0,\pi}(s_m,a_m)\xi_t(s_m,a_m) \\ & \leq \gamma \sqrt{\frac{4|S_{\mathcal{L}(s)+1}|  |\A| \ln \frac{3L T}{\delta}}{t}}.
\end{align*}
Then we have: 
 \begin{align*}&\sum_{t=1}^T\sum_{l=1}^L \sum_{m=0}^{l-1}\sum_{s_m \in \St_m}\sum_{a_m\in \A}q^{P_0,\pi}(s_m,a_m)\xi_t(s_m,a_m)  \\
& \leq \sum_{t=1}^TL^2\gamma \sqrt{\frac{4|\St|  |\A| \ln \frac{3L T}{\delta}}{t}}\\
& = L^2\gamma \sqrt{4T|\St|  |\A| \ln \frac{3L T}{\delta}}
\end{align*}
\end{proof}
\subsection{Proof of Theorem 2}
\begin{proof}[Proof of Theorem 2] 
From Lemma 2, 
\[
 v^{\pi^*}(s_0)-v^{\pi_t}(s_0)=\langle q^*-q_t, r\rangle\leq \|q^{P_0,\pi^*}-q^{P_t,\pi^*}\|_1. 
\]
From Lemma 3 and the proof of Lemma 5, we have the following holds with probability at least $1-\delta$: \begin{align*}
\|q^{P_0,\pi^*}-q^{P_t,\pi^*}\|_1 \leq  L^2\gamma \sqrt{\frac{4|\St|  |\A| \ln \frac{3L T}{\delta}}{t}}.
\end{align*}
When $t \ge \frac{4L^4 \gamma^2 |\St||\A| \ln \frac{3L T}{\delta}}{\epsilon^2} $, we have \(v^{\pi^*}(s_0)-v^{\pi_t}(s_0) \leq \epsilon\) with probability at least $1-\delta$.
\end{proof}
\subsection{Proof of Theorem 3}
\begin{proof}[Proof of Theorem 3]
When $t\ge  \lceil \frac{8|\St|^2|\A|\ln \frac{3LT}{\delta}}{h}\rceil$, for any $l$, $K_{l,t}=\{k\in K: \|P^k(s_{tl},a_{tl})- \hat{P}_t(s_{tl},a_{tl})\|_1\leq \sqrt{\frac{4|\St_{l+1}||\St_l||\A|\ln \frac{3LT}{\delta}}{t}}<\frac{1}{2}$h. 

With probability at least $1-\delta$, for all $l\in [L]$,
\begin{align*}
& \|P_0(\cdot |s_{tl},a_{tl})-\hat{P}_t(\cdot |s_{tl},a_{tl}) \|_1 \\
&\leq \sqrt{\frac{4|S_{\mathcal{L}(s)+1}|^2 |\A| \ln \frac{3LT }{\delta}}{t}}\\
&\leq \frac{1}{2}h.
\end{align*}
So for any $l\in [L]$, the true prototype, $k^*_l$, is in $K_{l,t}$. Then consider $k\not= k^*_l$, we have $\|P^k(\cdot |s_{tl},a_{tl})-\hat{P}_t(\cdot |s_{tl},a_{tl})\|_1\ge \|P^k(\cdot |s_{tl},a_{tl})-P_0(\cdot |s_{tl},a_{tl})\|_1-\|P_0(\cdot |s_{tl},a_{tl})-\hat{P}_t(\cdot |s_{tl},a_{tl})\|_1 > h-\frac{1}{2}h=\frac{1}{2}h$. Thus, for any $l$, $K_{tl}=\{k^*_l\}$. We have found the optimal prototype. 
\end{proof}

\section{Non-robust Algorithm and Theoretical Guarantees} \label{app:alg_nr}
\subsection{Non-robust Prototype Selection and Policy Update Algorithm}
\begin{algorithm}[ht]
\caption{Algorithm NRPO-NPC}
\begin{algorithmic}[1]
\REQUIRE State space $\mathcal{S}$, action space $\mathcal{A}$, episode number $T$, prototypes set $\otimes_{l\in [L]}\mathcal{K}_l$
\STATE \textbf{Initialization:} $\pi \leftarrow \pi_0$, number of samples $N_0(s,a)=0$ for each $s\in \mathcal{S}$, $a\in \mathcal{A}$
\FOR{$t = 1, \dots, T$}
    \FOR{$l = 1, \dots, L$}
        \STATE $s_{tl}, a_{tl} = \arg \max_{s \in \mathcal{S}_l, a \in \mathcal{A}} N(s,a)$
        \STATE Select the prototype that is closest to the empirical distribution: \\ 
        $k_l = \argmin_{k\in \mathcal{K}_{l}} \|P^k(s_{tl},a_{tl}) - \hat{P}_t(s_{tl},a_{tl})\|_1$
    \ENDFOR
    \STATE Construct the transition kernel: \\
    $P = \bigotimes_{l\in \mathcal{L},s\in \mathcal{S}_l,a\in \mathcal{A}} P^{k_l}(s,a)$
    \STATE Update policy $\pi_t = \max_{\pi}R(\pi, P)$
    \STATE Execute policy $\pi_t$ for $L$ steps and obtain trajectory $\mathcal{S}_l, a_l$ for $l = 1, \dots, L-1$
    \STATE Update $N_t(s,a)$ for all $s,a$ and the empirical distribution $\hat{P}_t(s,a)$ for all $s, a$
\ENDFOR
\end{algorithmic}
\end{algorithm}
% \nobibliography*
\subsection{Proof of Lemma 6}
\begin{proof}[Proof of Lemma 6]
From inequality \eqref{eq:1norm-dif}, we can bound the one-norm difference between $P_0$ and $\hat{P}_t$. 
\[
\|P_0(\cdot |s_{tl},a_{tl})-\hat{P}_t(\cdot |s_{tl},a_{tl}) \|_1\leq \sqrt{\frac{4|\St_{l+1}|\ln \frac{3LT}{\delta}}{\min\{N_{t}(s_{tl},a_{tl}),1\}}}
\]
Since we select the prototype that is most close to the empirical distribution $\hat{P}_t$,
\begin{align*}
&\|P^{k_t}(\cdot |s_{tl},a_{tl})-\hat{P}_t(\cdot |s_{tl},a_{tl}) \|_1\\
&\leq \|P_0(\cdot |s_{tl},a_{tl})-\hat{P}_t(\cdot |s_{tl},a_{tl}) \|_1 \\
&\leq \sqrt{\frac{4|\St_{l+1}|\ln \frac{3LT}{\delta}}{\min\{N_{t}(s_{tl},a_{tl}),1\}}}
\end{align*}
Thus, 
\[
\|P_0(\cdot |s_{tl},a_{tl})-P_t(\cdot |s_{tl},a_{tl}) \|_1\leq 4\sqrt{\frac{|\St_{l+1}|\ln \frac{3LT}{\delta}}{\min\{N_{t-1}(s_{tl},a_{tl}),1\}}}.
\]
Then from the proof of Lemma 4, we have:
\[
\|P^{k_t}(\cdot |s,a),P^k(\cdot |s,a) \|_1\leq \sqrt{\frac{4|S_{\mathcal{L}(s)+1}|  |\A| \ln \frac{3L T}{\delta}}{t}}
\] 
\end{proof}
Combining this lemma with the regret decomposition \eqref{eq:6}, Lemma 3 and 5, we have the same regret bound the robust algorithm. Then Theorem 2 follows for the non-robust algorithm. For the convergence of the prototype, from the proof of Theorem 3, \(\|P_0(\cdot |s_{tl},a_{tl})-P_t(\cdot |s_{tl},a_{tl}) \|_1\leq \frac{1}{2}h \),  \(\|P^k(\cdot |s_{tl},a_{tl})-P_t(\cdot |s_{tl},a_{tl}) \|_1> \frac{1}{2}h \) for any $k\not= k_l^*$, thus selecting the prototype that is closest to the empirical distribution yields the same result as in Theorem 3. 
\subsection{Algorithm NRPO-NPC2}
\begin{algorithm}[ht]
\caption{Algorithm NRPO-NPC2}
\begin{algorithmic}[1]
\REQUIRE State space $\mathcal{S}$, action space $\mathcal{A}$, episode number $T$, prototypes set $\otimes_{l\in [L]}\mathcal{K}_l$
\STATE \textbf{Initialization:} $\pi \leftarrow \pi_0$, number of samples $N_0(s,a)=0$ for each $s\in \mathcal{S}$, $a\in \mathcal{A}$
\FOR{$t = 1, \dots, T$}
   \FOR{$l = 1, \dots, L$}
       \STATE Select the prototype that is closest to the empirical distribution: \\
       $k_l = \argmin_{k\in \mathcal{K}_{l}} \sum_{s\in \mathcal{S}_l, a\in \mathcal{A}}\|P^k(s_{tl},a_{tl}) - \hat{P}_t(s_{tl},a_{tl})\|_1$
   \ENDFOR
   \STATE Construct the transition kernel: \\
   $P = \bigotimes_{l\in \mathcal{L},s\in \mathcal{S}_l,a\in \mathcal{A}} P^{k_l}(s,a)$
   \STATE Update policy $\pi_t = \max_{\pi}R(\pi, P)$
   \STATE Execute policy $\pi_t$ for $L$ steps and obtain trajectory $\mathcal{S}_l, a_l$ for $l = 1, \dots, L-1$
   \STATE Update $N_t(s,a)$ for all $s,a$ and the empirical distribution $\hat{P}_t(s,a)$ for all $s, a$
\ENDFOR
\end{algorithmic}
\end{algorithm}
In each episode, for each layer $l$, it selects the prototype $k_l$ that minimizes the sum of L1 distances between the prototype and empirical distributions across all state-action pairs in that layer. This differs from NRPO-NPC, which only considers the most sampled state-action pair. 

\end{document}